\documentclass[letterpaper, 10 pt, conference]{ieeeconf}  

\IEEEoverridecommandlockouts                              

\makeatletter
\let\NAT@parse\undefined
\makeatother
\usepackage[numbers]{natbib}  

\usepackage{amsmath} 
\usepackage{amssymb}  
\usepackage{booktabs}
\usepackage[ruled]{algorithm2e}
\usepackage{subfigure}
\usepackage{graphicx}
\usepackage{hyperref}

\title{\LARGE \bf
DARE: Diffusion Policy for Autonomous Robot Exploration
}

\author{Yuhong Cao$^{*1}$, Jeric Lew$^{*1}$, Jingsong Liang$^1$, Jin Cheng$^2$, Guillaume Sartoretti$^{1}$
\thanks{* The first two authors contributed equally to this work.}
\thanks{$^{1}$ Department of Mechanical Engineering, National University of Singapore, Singapore.
        {\tt\small \{caoyuhong, jsliang, mpegas\}@nus.edu.sg, jericlew@u.nus.edu}}%
\thanks{$^{2}$ Department of Computer Science, ETH Z{\"u}rich, Switzerland.
    {\tt\small jicheng@ethz.ch}}%
}

\begin{document}

\maketitle
\thispagestyle{empty}
\pagestyle{empty}

\begin{abstract}

Autonomous robot exploration requires a robot to efficiently explore and map unknown environments. Compared to conventional methods that can only optimize paths based on the current robot belief, learning-based methods show the potential to achieve improved performance by drawing on past experiences to reason about unknown areas. In this paper, we propose DARE, a novel generative approach that leverages diffusion models trained on expert demonstrations, which can explicitly generate an exploration path through one-time inference. We build DARE upon an attention-based encoder and a diffusion policy model, and introduce ground truth optimal demonstrations for training to learn better patterns for exploration. The trained planner can reason about the partial belief to recognize the potential structure in unknown areas and consider these areas during path planning. Our experiments demonstrate that DARE achieves on-par performance with both conventional and learning-based state-of-the-art exploration planners, as well as good generalizability in both simulations and real-life scenarios.

\end{abstract}

\section{INTRODUCTION}

In autonomous robotic exploration, a mobile robot is tasked with fully mapping all accessible parts of an unknown environment as fast as possible. Typically, the robot is equipped with sensors, such as a 360-degree Lidar, to gather geometric information, allowing it to incrementally construct a more complete belief (e.g., a 2D occupancy map) of the environment.
\textit{Frontiers} are cells on the boundary between free and unknown areas in the robot belief, and are widely used to represent the information in exploration and drive the planning~\cite{yamauchi1997frontier,bircher2016receding,dang2020graph,cao2021tare,cao2023ariadne}. 
In this work, we study innovative frontier-driven exploration methods that consider localization and mapping to be nearly perfect thanks to the recent advancements in Lidar odometry~\cite{xu2021fast,bai2022faster}.

State-of-the-art conventional robot exploration planners rely on well-designed optimization algorithms that aim to find the shortest path to observe the most unexplored areas. To find such a path, an exploration planner needs to reason about the full robot belief for decision-making. Additionally, since the robot belief will be updated frequently during the exploration, the planner must be computational-efficient to reactively handle the belief changes as well. Such requirements highlight the importance of balancing the trade-off between optimization quality and computing time in exploration planning. Most recent advanced exploration planners were developed on sampling-based strategies~\cite{bircher2016receding,dang2020graph,cao2021tare,huang2023fael} for their outstanding efficiency in exploration path planning. 

\begin{figure}[t]
    \vspace{+0.2cm}
    \centering
    \includegraphics[width=0.9\linewidth]{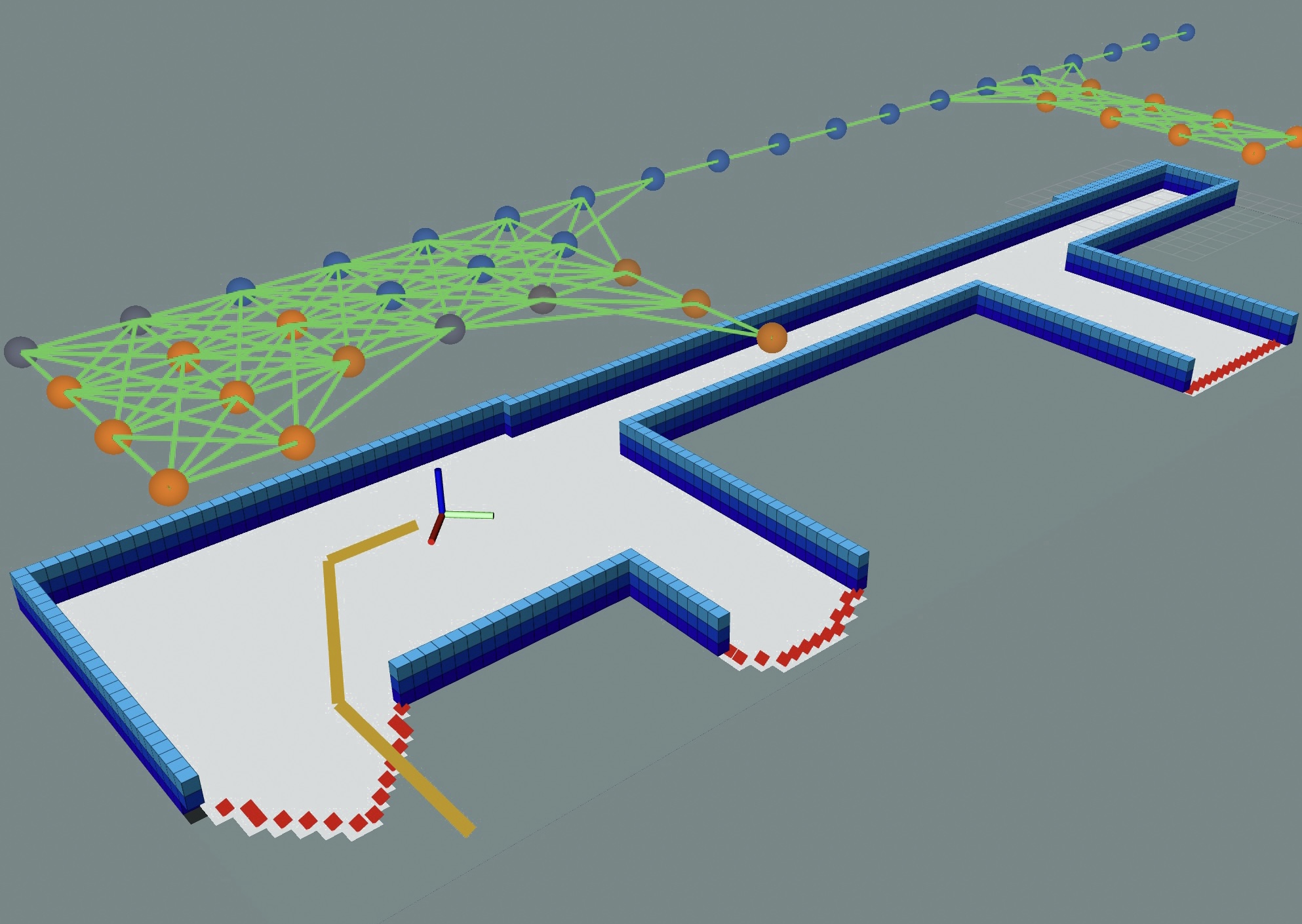}
    \caption{\textbf{An example planned path from DARE.} Based on the robot belief (represented by the occupancy grid map), a robot (represented by the axes) constructed an informative graph. The graph is passed to an attention encoder network and a diffusion policy network to output a planned exploration path (in orange). The robot executes the planned path in the receding horizon manner until it fully explores the environment. Note DARE can reason about the partial belief to recognize the potential structure in some unknown areas and consider these areas during path planning. }
    \label{fig: example}
    \vspace{-0.6cm}
\end{figure}

Since exploration tasks challenge planners to rapidly reason about the long-term efficiency of planned paths, some recent works have looked to deep reinforcement learning (DRL) to tackle exploration due to its ability to make reactive decisions and estimate long-term returns. 
Notably, our previous works~\cite{cao2023ariadne,cao2024deep} built upon attention-based networks~\cite{vaswani2017attention} achieves on-par or even better performance than the state-of-the-art conventional planners, as well as robust generalizability to real-life developments. We believe that the exciting performance of DRL planners comes from their implicit predictions of the future robot belief. Similarly to humans, learning-based planners can draw on past experiences to reason about what might lie in unexplored areas. However, most DRL-based planners typically only generate the next waypoint, lacking visible and explainable long-term planning. In more challenging exploration scenarios, environments may contain long corridors and deadends resulting in sparsity in reward signals that makes training difficult. 

To further study the application of advanced learning-based technology in robot exploration and expand the research horizon of the community, we propose DARE, abbreviated for Diffusion Policy for Autonomous Robot Exploration, a novel generative approach that leverages diffusion models trained on expert demonstrations. DARE not only generates explicit long-term paths, but also achieves improved performances over state-of-the-art DRL-based planners in challenging exploration scenarios. Although DARE is trained using supervised learning/behavior cloning, which is usually less effective than deep reinforcement learning in policy training, we address this disadvantage by introducing a ground truth optimal demonstration planner. Specifically, we build upon the observation that exploration can be solved optimally in known environments by planning a complete coverage, and give our expert planner access to the ground truth map and build training datasets with optimal exploration paths. As a result, our trained planner can reason about the partial belief to recognize the potential structure in some unknown areas and consider these areas during path planning. Instead of directly encoding the robot belief, we construct an informative graph from the robot belief and encode it as the input of diffusion networks, providing better exploration performance and generalizability. To validate DARE, we conduct comparison experiments with a wide range of learning and conventional baseline methods, as well as demonstrations in a Gazebo simulation and a real-world deployment.  

\section{RELATED WORKS}

\subsection{Conventional Exploration Planners}
We refer to methods that select one of the frontiers as the navigation target for exploration as frontier-based methods. Frontier-based exploration methods have been widely studied for decades and a range of greedy strategies were developed to improve the exploration performance in the relatively simple exploration scenarios~\cite{yamauchi1997frontier,gonzalez2002navigation,kulich2011distance,holz2010evaluating}. For these frontier-based methods, the selection of the target frontier could be based on \textit{cost} (i.e., the distance to a frontier), \textit{utility} (i.e., observable information at a frontier), or their combinations~\cite{holz2010evaluating,kulich2011distance,meng2017two,cieslewski2017rapid,osswald2016speeding}. More recent works~\cite{bircher2016receding,dang2020graph,cao2021tare,huang2023fael,zhu2021dsvp} significantly improve the exploration efficiency by planning non-myopic exploration paths that consider covering more frontiers. These methods typically rely on well-designed sampling strategies to balance the trade-off between planning quality and computation time to generate long-term exploration paths and execute them in a receding-horizon manner.

\subsection{Learning-based Exploration Planners}
Recently, deep reinforcement learning-based exploration methods have been introduced to tackle exploration problems by training a policy to maximize long-term returns. A range of DRL planners~\cite{zhu2018deep,li2019deep,xu2022explore,chen2020autonomous} based on convolutional neural networks (CNNs) were proposed and validated in the small-scale robot exploration tasks. However, CNN-based exploration planners are limited to small-scale environments due to the predefined size of network input. Meanwhile, the performance of trained CNN-based policies is typically poorer than advanced conventional methods~\cite{cao2023ariadne}. On the other hand, our previous works~\cite{cao2023ariadne,cao2024deep} proposed to construct a graph representation of the robot belief as the input to attention-based graph neural networks. Leveraging the ability of the attention mechanism~\cite{vaswani2017attention} to capture long-term dependencies, attention-based DRL planners show on-par or even better performance than advanced conventional planners and good generalizability to real-life applications.   

\subsection{Diffusion Model in Robotics}
\label{sec:diffusion_related_work}
Diffusion models~\cite{sohl2015deep,ho2020denoisingdiffusionprobabilisticmodels} are originally developed for image generation tasks and have demonstrated state-of-the-art results in many generative tasks. Diffusion models aim to model the process of gradually adding Gaussian noise to data, and train neural networks to reverse this process to generate new data.
To leverage the strength of diffusion models to boost robot learning, recent works~\cite{janner2022diffuser,chi2023diffusionpolicy,chi2024diffusionpolicy,sridhar2023nomad, yu2024ldplocaldiffusionplanner} have applied diffusion models to robotic tasks such as motion planning for robotic manipulators, visual navigation and collision avoidance. These diffusion-based planners rely on the \textit{state} of the robot to generate appropriate action sequences. This \textit{state} can be represented either by raw data, such as pose information, or by encoded observations. Typically, RGB images are processed through visual encoders, such as ResNets~\cite{He_2016_CVPR}, to obtain state representations in a latent feature space. Using the robot’s \textit{state}, these diffusion models then sample a sequence of actions from Gaussian noise and iteratively denoise it to produce a noise-free sequence of actions.
To train diffusion models for robotic tasks, existing works rely on offline reinforcement learning~\cite{janner2022diffuser} or behavior cloning~\cite{chi2023diffusionpolicy, chi2024diffusionpolicy}.
There, the high-dimensional output space of diffusion models allows them to generate a sequence of future actions rather than individual actions in succession. This promotes temporally consistent actions, allowing the robot to understand and model how its current actions will affect its future actions.
Notably, diffusion policy~\cite{chi2023diffusionpolicy, chi2024diffusionpolicy} models the conditional action distribution instead of the joint state-action distribution \cite{janner2022diffuser}, which eliminates the need to include the observation in every denoising iteration. This allows for end-to-end training of the diffusion model and observation encoder (for conditioning of the diffusion process) while also significantly improving inference speeds, allowing real time deployment, a key requirement for autonomous exploration.

\section{PROBLEM STATEMENT}

Given an environment \(\mathcal{E}\) represented by a 2D occupancy grid map \(M\), where \(M\) consists of free space \(M_f\) and occupied space \(M_o\), such that \(M = M_f \cup M_o\). A robot within this environment maintains a 2D occupancy grid map \(B\), representing the robot's belief about \(\mathcal{E}\). The map \(B\) consists of unknown space \(B_u\), free space \(B_f\), and occupied space \(B_o\), where \(B = B_u \cup B_f \cup B_o\). The autonomous exploration problem begins with no prior knowledge of the environment, meaning \(B = B_u\). As the robot explores \(\mathcal{E}\) using a sensor, the area of the environment perceived at each step is determined by the sensor's measurements and is denoted as \(M_s\), where \(M_s \subset M\). The robot's belief map \(B\) is then updated with the newly explored regions \(M_s\). 
The exploration is considered complete when $B_f = M_f$, which in practice is equivalent to the close of occupied space $M_o$.
The goal of autonomous exploration is to complete the exploration with the shortest possible collision-free trajectory \(\psi^*\) given by:
\begin{equation}
\psi^* = \underset{\psi \in \Psi}{\arg\min} \, \text{C}(\psi), \quad \text{s.t.} \quad B_f = M_f
\end{equation}
where $\text{C}: \psi \rightarrow \mathbb{R}^+$ maps a trajectory to its length. 

\begin{figure*}
  \centering
  \includegraphics[width=0.99\linewidth]{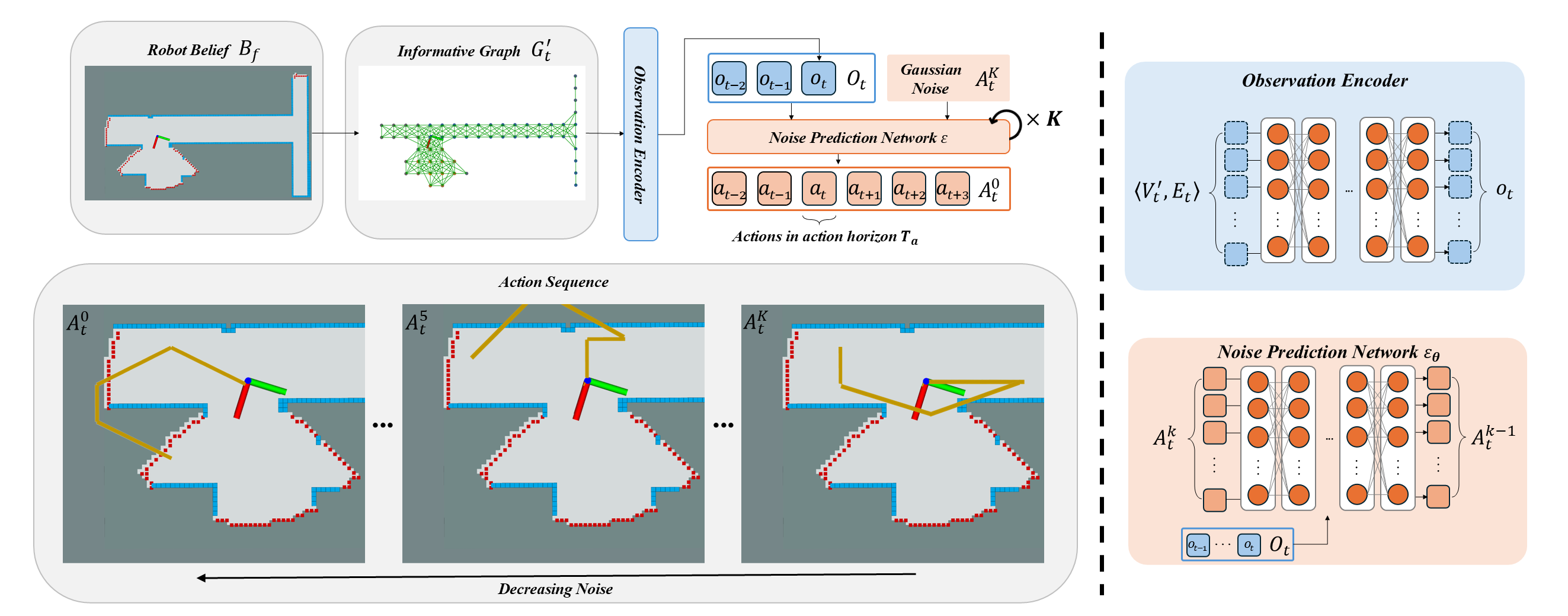}
  \vspace{-10pt}
  \caption{\textbf{Diffusion-based exploration planner.} At each step, DARE maintains a graph-based belief representation and encodes it with self-attention layers to capture a robot belief feature. Conditioned on a sequence of robot belief features, the diffusion policy generates future action sequences through iterative denoising. Note that planned paths can extend to unknown areas.
  }
  \label{fig: workflow}
  \vspace{-10pt}
\end{figure*}

\section{METHODOLOGY}
To tackle the autonomous exploration problem, we propose DARE, a diffusion-model-based exploration planner, and train it to plan efficient exploration paths. Our method builds upon our previous works~\cite{cao2023ariadne,cao2024deep}, which formulate exploration as a path planning problem on a graph and encode the robot belief with attention-based networks. In particular, we first construct an informative collision-free graph from the partial map $B$ to represent the robot belief. This graph is then encoded using an attention-based encoder as a robot belief feature in latent space. Conditioned on this encoded belief feature, we utilize a diffusion policy network to generate a multi-step trajectory and execute it in a receding horizon manner. Notably, we further propose a ground truth optimal exploration path planner to provide demonstrations for training our diffusion policy. Since this planner is only used to collect training data, we allow it to access the ground truth map instead of the partial belief map only. Consequently, it is easy for this sampling-based planner to find the optimal (or at least near-optimal) coverage path, which is approximately equivalent to the optimal exploration path. By doing so, our diffusion-based planner is targeted to reproduce such near-optimal exploration paths while providing generalizability.       

\subsection{Graph-based Exploration}
Graph representations of environments have been widely used for exploration planning as they efficiently discretize the action space and explicitly represent collision-free connections between nodes through edges~\cite{dang2020graph,cao2021tare,cao2023ariadne}. 
At each decision step $t$, we construct a collision-free graph with uniformly distributed nodes, 
\( V_t = \{v_0, v_1, \dots\}, \forall\ v_i = (x_i, y_i) \in B_f\} \) in the currently observed free space $B_f$, each at a distance of $d_n$ meters (node resolution). Each node is connected to up to $5\times5$ nearest neighboring nodes around itself, where connections are made only if a collision-free straight-line path exists between them. Consequently, we have a collision-free graph $G_t = (V_t, E_t)$, where $V_t$ is the set of uniformly distributed nodes in the free space, and $E_t$ is the set of edges representing collision-free paths between nodes. These nodes serve as candidate viewpoints/waypoints for the robot, where one of them $v_t$ is at the robot's current position. To get collision-free paths, we only allow the robot to move on valid edges (i.e., to one of neighboring nodes). 
Thus, the planned exploration path is a sequence of future locations $\psi_{t+1:t+n} = [\psi_{t+1}, \psi_{t+2}, ..., \psi_{t+n}]$, where $\forall \psi_{i} \in V$ and $n$ denotes the length of the planned trajectory. We let the robot execute the planned path in the receding horizon manner, where the robot executes the first $T_a$-step of the planned trajectory and then replans the exploration path. 

\subsection{Observation Encoder}

To enable the diffusion policy to accurately perceive the current state, it is conditioned on a learned high-dimensional feature, typically encoded features from images through a visual encoder as mentioned in Section~\ref{sec:diffusion_related_work}. In the context of our exploration problem, since the robot maintains a 2D occupancy map, training a diffusion policy conditioned on the map images naturally comes to our mind. However, we note that there are some critical drawbacks of learning diffusion policy from images. First, visual encoders can not process images with larger sizes than the training ones, but an unknown environment could be in any size and shape, resulting in a poor generalizability of the trained policy. Second, CNN-based encoders are not good at capturing long-term dependencies within the exploration space, which has been experimentally validated in our previous work~\cite{cao2023ariadne}. 
Last but not least, in real-life applications, the occupancy map could be cluttered and noisy, making it drastically different from the training maps and leading to a severe sim-to-real gap.

\noindent \textbf{Graph Observation:} On the other hand, encoding graph representations with attention-based networks has been validated in our previous works~\cite{cao2023ariadne, cao2023catnipp} as an efficient approach to learning robot belief for exploration planning, with no constraint on the map shape/size and robust sim-to-real generalizability~\cite{cao2024deep}. Therefore, we take the attention-based network in~\cite{cao2023ariadne} as the backbone of the encoder for our diffusion policy. Similar to~\cite{cao2023ariadne}, we construct an \textit{informative graph} $G'_t=(V'_t, E_t)$ from the robot belief and the collision-free graph $G_t$. This graph not only provides traversable trajectory space for path planning but also contains the information distribution (i.e., frontiers) in the robot belief. Note that $G'_t$ shares the same edge set $E_t$ as $G_t$. In addition to the node coordinates (i.e., $v_i=(x_i, y_i)$), The properties of each node $v'_i$ in the informative graph further include the \textit{utility}, which quantifies the information gain at node $v_i$ through the number of observable frontiers, the \textit{guidepost}, which is a binary signal $b_i$ indicates paths to nodes with non-zero utility, such that $v'_i=(x_i, y_i, u_i, b_i)$ and the \textit{occupancy}, which is a binary signal indicating the presence of the robot at the node. 

\noindent \textbf{Attention Encoder:} We project $V'_t$ to $d$-dimension features and pass it to $6$ stacked self-attention layers, where each attention layer takes the output of its previous layer as input. Each attention layer reads:
\begin{equation}
\begin{aligned}
& q_{i}=W^Qh^{q}_{i}, \ k_{i}=W^Kh^{k,v}_{i}, \ v_{i}=W^Vh^{k,v}_{i}, \ u_{ij}=\frac{q_{i}^T\cdot k_{j}}{\sqrt{d}}, \\ & w_{ij}=\left\{\begin{array}{cc}
    \frac{e^{u_{ij}}}{\sum_{j=1}^{n}e^{u_{ij}}}, & M_{ij}=0 \\
    0, & M_{ij}=1
\end{array}\right., \ h'_{i}=\sum_{j=1}^{n}w_{ij}v_{j}, 1\leq i \leq m
\end{aligned}
\label{eq:attention}
\vspace{-0.1cm}
\end{equation}
where each $h_i\in\mathbb{R}^{d\times 1}$ is a $d$-dimension feature vector associated with node $v'_i$, superscripts $q$ and $k,v$ denote the source of the \textit{query}, \textit{key}, and \textit{value} respectively~\cite{vaswani2017attention}, $W^Q, W^K, W^V \in \mathbb{R}^{d\times d}$ are learnable matrices, and $M$, which serves as an edge \textit{mask}, is a $m\times m$ adjacency matrix ($m$ denotes the number of nodes) built from $E$ ($M_{ij}=0$, if $(i, j) \in E$, else $M_{ij}=1$). This edge mask constrains $h_i$ to access features of its neighboring nodes to model the graph connectivity.
The output of self-attention layers is a group of \textit{node features}, each $h'_i$ modeling dependencies between node $v'_i$ and all other nodes. We term the node feature associated with the current node $v_t$ as the \textit{current node feature}. We then pass the current node feature and all node features to a cross-attention layer, where the structure is the same as self-attention layers but the query source is the current node feature only and the key-value sources are all node features. 
This cross-attention layer merges all node features with the current node feature, encapsulating the robot's understanding of the environment, thus we term it the \textit{robot belief feature}. This feature represents the robot belief in the latent space and serves as an input to the diffusion policy network.

\subsection{Diffusion Policy}

Our method builds upon diffusion policy~\cite{chi2023diffusionpolicy, chi2024diffusionpolicy}, which uses a supervised behavior cloning approach and consistently outperforms previous state-of-the-art methods in robotic manipulator behavior cloning benchmarks. 
To generate an action sequence at timestep $t$, the diffusion policy begins with an initial noisy action sequence $A_t^K$ drawn from a Gaussian distribution. $A_t^K$ is then iteratively denoised over $K$ iterations by subtracting the output of a noise-prediction model $\varepsilon_\theta$, resulting in $A_t^0$, a denoised action sequence. The denoising process is described by:
\begin{equation}
A_t^{k-1} = \alpha \left( A_t^k - \gamma \varepsilon_\theta (O_t, A_t^k, k) + \mathcal{N}(0, \sigma^2 \mathbb{I}) \right),
\end{equation}
where $O_t$ is the observation condition (\textit{robot belief feature}), $k$ is the current denoising step, and $\mathcal{N}(0, \sigma^2 I)$ is Gaussian noise. The parameters $\alpha$, $\gamma$, and $\sigma$ are functions of $k$ and define the noise schedule, which is analogous to the learning rate schedule in stochastic gradient descent.
At time step $t$ and denoising iteration $k$, the action sequence $A_t^k$ represents a series of action $a_{t+i}^k$ such that $A_t^k = [a_{t-T_o+1}^k, \dots ,a_{t}^k, \dots , a_{t-T_o+T_p}^k]$ where $T_o$ is the observation horizon and $T_p$ is the prediction horizon. Only actions within the action horizon $T_a$, given as $[a_{t}^k, a_{t+1}^k,..., a_{t+T_a-1}^k]$ is executed in a receding horizon manner.

In our graph-based exploration, actions are defined as a selection of one of the aforementioned 5x5 neighboring nodes (including itself) represented as a one-hot encoding. Specifically, each action $a_{t,i}^k$ is a 1D vector of 10 binary values: the first 5 indices represent the x-index and the next 5 represent the y-index of the selected neighboring node. When a neighboring node is chosen, we check if an edge exists in $G_t$ that connects the current node to the selected neighbor, ensuring a collision-free path.
To obtain the planned exploration path $\psi_{t+1:t+n}$, we first discard past actions $a_i^k$ for $i < t$. Then, we map the remaining one-hot encoding of the denoised action $A_t^0$ to the positional change of the corresponding neighboring node, denoted as ${A'}_t^0$, by leveraging the uniform structure of the collision-free graph with a node resolution (i.e., the gap between nodes) $d_n$. Finally, we cumulatively sum ${A'}_t^0$ with respect to the robot’s current location to generate the sequence of future locations $\psi_{t+1:t+n}$ where $n = T_p - T_o + 1$. Only $T_a$, the action horizon, number of steps within the sequence of future locations will be executed to promote reactivity to new information during exploration. Note that we utilize the transformer-based noise-prediction network.
We train the noise-prediction network $\varepsilon_\theta$ with mean squared error (MSE) loss which reads:
\begin{equation}
\vspace{-0.1cm}
\mathcal{L} = \mathbb{E}[(\varepsilon^k - \varepsilon_\theta \left(O_t, A_t^0 + \varepsilon^k \right))^2],
\label{eq:loss_function}
\end{equation}
where $\varepsilon^k$ is a sampled random noise with appropriate variance and $A_t^0$ is the noise-free action sequence from expert demonstration. We note that the generated paths may have collisions, thus we further implemented a simple collision avoidance algorithm to ensure that at least the part within the action horizon is collision-free.

\begin{table*}[t]
\caption{
\textbf{Comparisons with baseline planners (identical 100 scenarios for each method).} The travel distance lower the better.}
\label{table:1}
\begin{center}
\begin{tabular}{c|cccc|cc|c}
\toprule
& Nearest & Utility & NBVP & TARE Local & ARiANDE & Ours & Optimal\\
\midrule
Distance (m) & 652($\pm$76) & 585($\pm$79) & 645($\pm$109) & 558($\pm$67) & \ 579($\pm$82) & \ 563($\pm71$\ ) & 499($\pm61$) \\
\midrule
Gap to Optimal & 30.6\% & 17.2\% & 29.2\% & 11.8\% & 16.0\% & 12.8\% &  0\% \\
\bottomrule
\end{tabular}
\end{center}
\vspace{-0.5cm}
\end{table*}

\subsection{Ground Truth Optimal Demonstration}

The diffusion policy is only trained to clone the expert's policy, thus the trained performance of the diffusion policy can never surpass the performance of the expert planner. As a result, if we take any existing exploration planner as the expert, it can be foreseen that we can only get a sub-optimal diffusion policy for exploration. To address this issue, we let the expert planner have access to the ground truth map of the environment, thus the exploration problem is simplified to the coverage problem, and the optimal coverage path is identical to the optimal exploration path.
Even though it may be impossible to find the optimal coverage/exploration paths from only a partial map, we train our diffusion model to capture useful underlying patterns by reasoning about the dependencies between the optimal exploration path and the partial robot belief, leading to high-quality planned paths.  

In this work, we extend the constrained sampling-based planner proposed in~\cite{cao2021tare} to find the shortest path for covering a known environment. We construct a ground truth graph $G^*(V^*, E^*)$, which is a collision-free graph that covers ground truth free area $M_f$. There, we consider the boundary of unexplored free area ($M_f-B_f$) as the \textit{ground truth frontiers} to be observed by the robot. Based on $G^*$, we sample a set of nodes with the constraint that the robot can cover all ground truth frontiers by visiting these nodes. Then we find the the shortest path to visit all these nodes by solving a Traveling Salesmen Problem (TSP)~\cite{ortools}. By iterating the above sampling process $k$ times, we select the shortest path among the sampled paths as the expert demonstrations. The algorithm is presented in Algorithm~\ref{alg: ground truth planner}.

\begin{algorithm}[t]
\caption{Ground Truth Optimal Planner}
\label{alg: ground truth planner}
\KwIn {Graph $G^*(V^*, E^*)$, robot position $p_t$, ground truth frontiers $F$}
\KwOut {Coverage path $\psi_c$}
Get observable frontiers $F_i$ for each node $v^*_i$ \\
Initialize priority queue $Q$ \\
$\forall v^*_i\in V^*$, push $v^*_i$ into $Q$ with priority as $u_i=|F_i|$ \\
$C_{best} \gets +\infty$, $\psi_c \gets \emptyset$\\
\For{$1 \ \mathrm{to} \ k$}{
    $Q' \gets Q$, $W \gets \{p_t\}$ \\ 
    \While {$Q'\neq \emptyset \ \mathrm{and}\ \exists u_j>0$}{
        Probabilistically sample $ v^*_k \sim Q'$\\
        $Q' \gets Q'/v^*_k$, $W \gets W\cup v^*_k$, $F\gets F/F_k$ \\
        Update $F_i$ and $u_i$ for all nodes in $Q'$ \\
        }
    Compute the TSP path $\psi$ over $W$ starting from $p_t$ \\
    \If {$C(\psi)<C_{best}$}{
        $C_{best} \gets C(\psi)$, $\psi_c \gets \psi$
    }
}
\end{algorithm}

\subsection{Training Details} 
We train DARE using simulated environments generated by a random dungeon generator, similar to~\cite{cao2023ariadne, chen2019selflearning} but with more challenging maps. Each environment is represented as a $250 \times 250$ grid map, corresponding to a $100 \text{m} \times 100 \text{m}$ area. The sensor range is set to $d_s = 20\text{m}$, and the node resolution is $d_n = 4\text{m}$. For training, we first generate a dataset of 4000 expert trajectories using our ground truth coverage planner.
DARE is trained to plan paths with planning horizon $T_p = 8$ based on observation horizon $T_o = 2$ and execute the action within action horizon $T_a = 1$. 
We use the Square Cosine Noise Scheduler \cite{nichol2021improved} with $K = 100$ denoising steps for the diffusion process. DARE is trained for 130 epochs, each consisting of 1158 iterations with a batch size of 256, using the AdamW optimizer \cite{loshchilov2018decoupled} at a learning rate of $10^{-4}$. We used 4 NVIDIA GeForce RTX 3090 GPUs for training and the training converged after 36 hours.

\section{EXPERIMENTS}

\begin{figure}[t]
    \centering
    \subfigure[Rooms]{\includegraphics[width=0.312\linewidth]{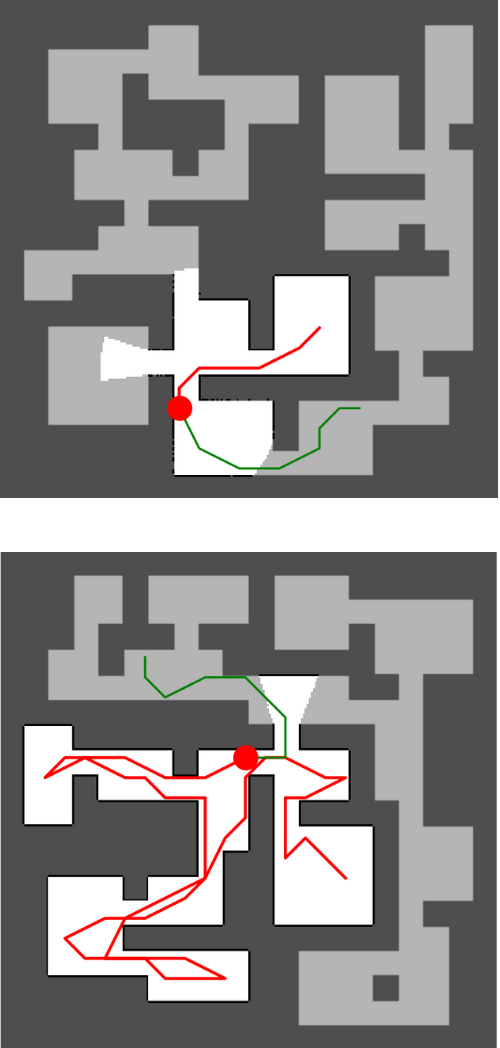}
        \label{fig:predicted_room}}
    \hspace{-0.5cm}
    \hfill
    \subfigure[Connections]{\includegraphics[width=0.312\linewidth]{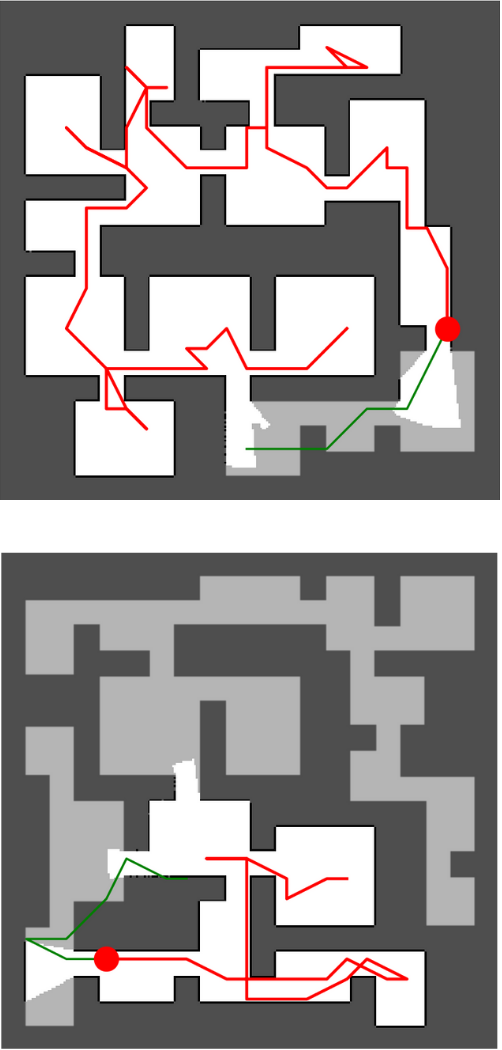}
        \label{fig:predicted_connections}}
    \hspace{-0.5cm}
    \hfill
    \subfigure[Deadends]{\includegraphics[width=0.312\linewidth]{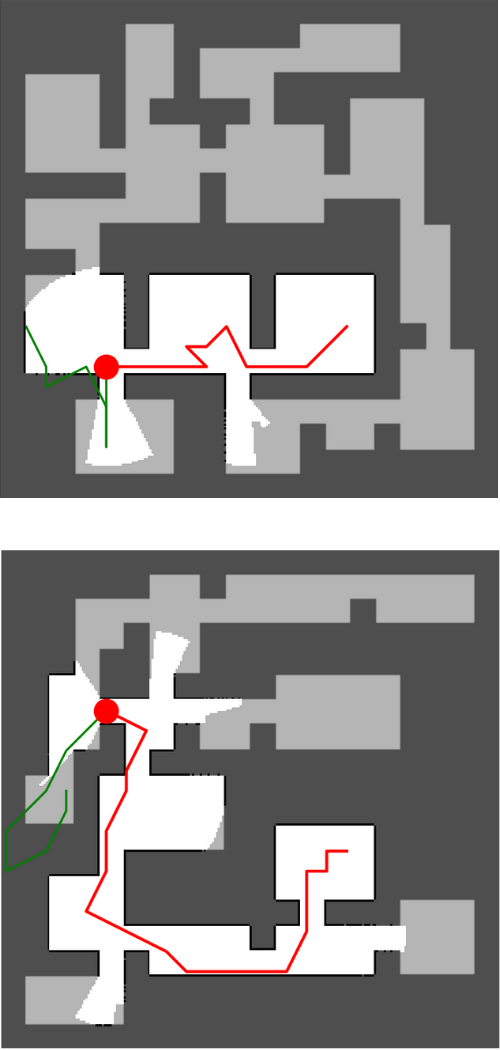}
        \label{fig:predicted_deadend}}
    \caption{\textbf{DARE exhibits the ability to predict unknown areas.} Here we show some examples where the predictions are correct. The previous trajectory of the robot (represented by the red dot) is in red. The planned path by DARE is in green. The explored free areas are in white while the unexplored free areas are in light gray.}
    \label{fig:predicted_paths}
    \vspace{-0.4cm}
\end{figure}

\subsection{Comparison with Baselines}
We conducted comparison experiments using a test set consisting of 100 unseen simulated dungeon environments. 
We compared the travel distance of DARE to complete the exploration with several conventional planners: Nearest~\cite{yamauchi1997frontier}, Utility~\cite{gonzalez2002navigation}, NBVP~\cite{bircher2016receding}, and TARE Local \cite{cao2021tare}. The Nearest method directs the robot to the closest frontier, while the Utility method evaluates the utility of frontier regions and the cost to reach them, with a tunable parameter $\lambda$ (0.1 in practice) to balance utility and cost. NBVP also randomly samples paths to determine the optimal viewpoint. TARE Local, the local planner of TARE \cite{cao2021tare}, plans the shortest trajectory that can cover all frontiers in the current partial map. Besides, we compare DARE to our previous work ARiADNE~\cite{cao2023ariadne}, a state-of-the-art graph-based DRL planner. We also provide the results of near-optimal paths generated by our ground truth coverage planner.

We define the exploration as completed when the occupied space is closed. The mean and variance of the trajectory lengths required to complete exploration are summarized in Table~\ref{table:1}. DARE outperforms other advanced conventional and learning baselines in terms of travel distance and achieves on-par performance with TARE Local, the state-of-the-art planner. Furthermore, we find that DARE surpasses TARE in $54\%$ of the test environments. To further support our claim on the on-par performance with TARE, we conduct a one-tailed paired t-test on the hypothesis that DARE's travel distance is significantly higher than TARE's. As a result, we get the p-value at $0.16$, much higher than the significance level of $0.05$. Therefore the hypothesis is rejected statistically, proving that DARE's performance is not significantly worse than TARE's in this test set.  

More than the numerical results, upon closer observations on planned paths from DARE, we find DARE exhibits a promising ability to predict unknown areas based on partial belief. We show some examples in Figure~\ref{fig:predicted_paths} where DARE accurately modeled the underlying structure in unknown areas and planned efficient exploration paths. These examples demonstrate the novelty and advantage of training the diffusion model using ground truth coverage demonstrations: the trained planner can reason about the partial belief to recognize the potential structure in some unknown areas and consider these areas during path planning. Although DARE's predictions are not perfect and could be even wrong, we believe DARE's performance will draw interest in studying the applications of diffusion models in robot exploration.

\begin{figure}
  \centering
  \includegraphics[width=0.99\linewidth]{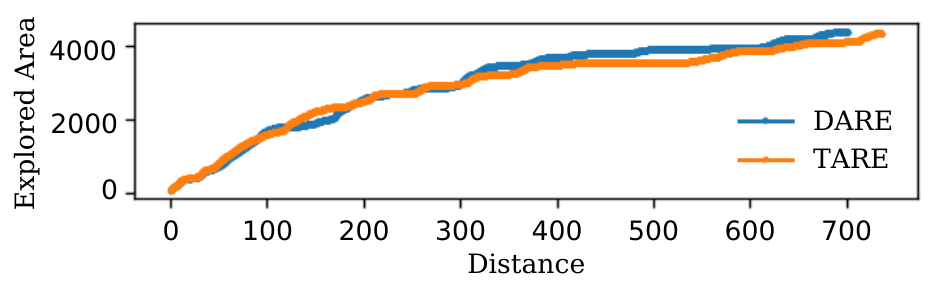}
  \vspace{-20pt}
  \caption{\textbf{Trajectory analysis in the Gazebo simulation.}}
  \label{fig: gazebo_curve}
  \vspace{-0.3cm}
\end{figure}

\begin{figure}[t]
    \centering
    \hfill
    \subfigure[TARE, $714\text{m}$, $736\text{s}$]{\includegraphics[width=0.225\textwidth]{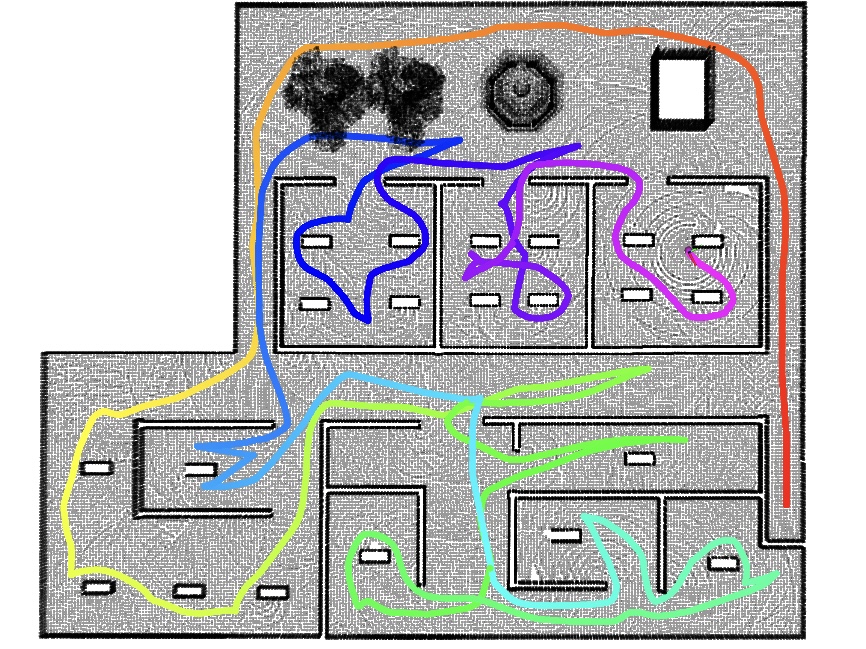}}
    \hfill
    \subfigure[DARE, $678\text{m}$, $701\text{s}$]{\includegraphics[width=0.225\textwidth]{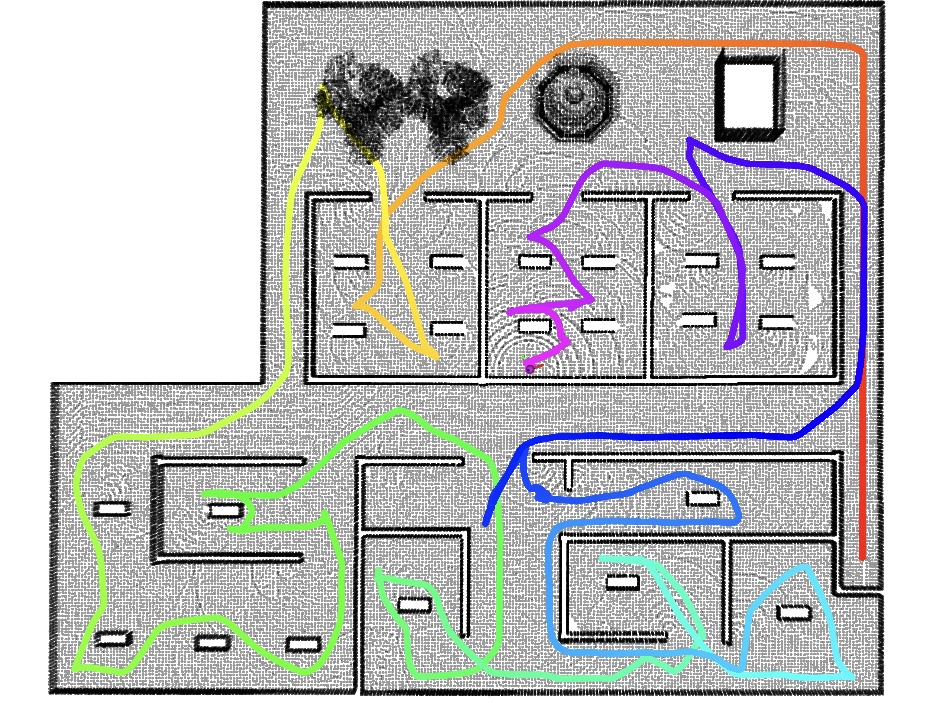}}
    \hfill
    \vspace{-0.1cm}
    \caption{
    \textbf{Exploration paths comparisons in the Gazebo simulation.}}
    \label{fig: gazebo}
    \vspace{-0.5cm}
\end{figure}

\subsection{Gazebo Simulation}

To further validate the effectiveness of DARE, we develop DARE in ROS and conduct a high-fidelity simulation in a $64\text{m}\times85\text{m}$ Gazebo environment provided by~\cite{huang2023fael}. In particular, we set the robot max speed to $1\text{m/s}$, the map resolution to $0.4\text{m}$, the node resolution to $2.8\text{m}$, and the sensor range to $20\text{m}$. 
We run the simulations on an ASUS PN53 mini PC with AMD R9-6900HX CPU (which is also used for our test on a real robot), where per-step planning time of DARE is on average $1.29\text{s}$ using a single-threading CPU process. As shown in Figure~\ref{fig: gazebo}, our planner completed the exploration in $701\text{s}$ and $678\text{m}$. In comparison, TARE completed the exploration in $736\text{s}$ and $714\text{m}$. 
Since we map the environment using Octomap while TARE uses point cloud, these two planners may not work in exactly the same condition. However, considering the completeness of our explored map in point cloud representation, we believe it is fair to say DARE achieves better exploration efficiency than TARE in this Gazebo environment.


\subsection{Hardware Validation}

\begin{figure}[t]
    \centering
    \hfill
    \subfigure[Mobile robot]{\includegraphics[width=0.16\textwidth]{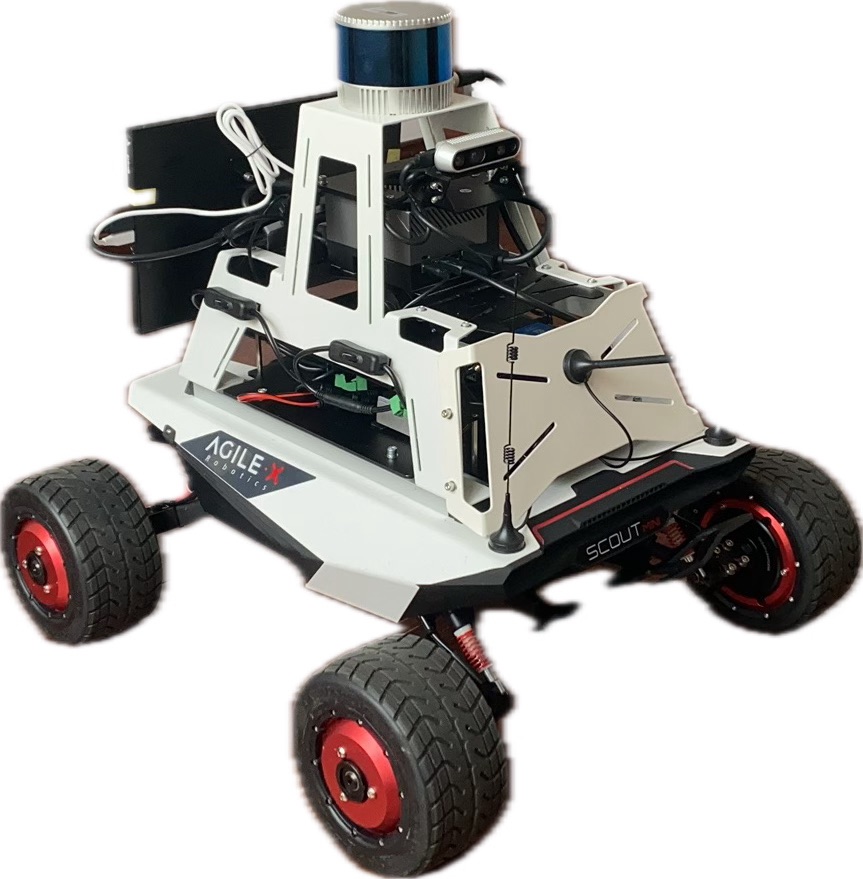}}
    \hfill
    \subfigure[Exploration path in a research lab]{\includegraphics[width=0.31\textwidth]{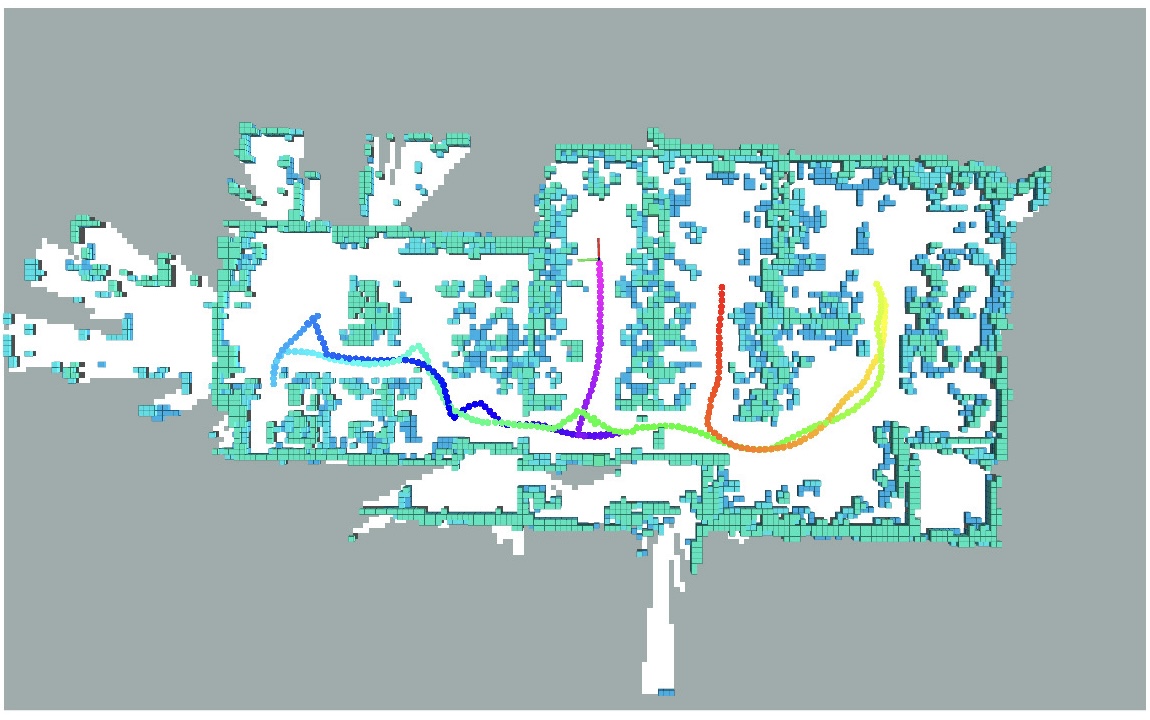}}
    \hfill
    \vspace{-0.1cm}
    \caption{\textbf{DARE guides a robot to explore a research lab.}}
    \label{fig: hardware}
    \vspace{-0.4cm}
\end{figure}

We further run our ROS planner on an Agilex Scout mini mobile robot equipped with a Leishen 16-line 3D lidar. The experiment is conducted in a $20\text{m}\times 30\text{m}$ research lab with cluttered furniture. In particular, we set the robot max speed to $0.5\text{m/s}$, the map resolution to $0.2\text{m}$, the node resolution to $1.6\text{m}$, and the sensor range to $10\text{m}$. As shown in Figure~\ref{fig: hardware}, guided by DARE, the robot completes the task of exploring the lab. We believe our hardware experiment validates that DARE is applicable to real-life applications.


\section{CONCLUSIONS}
In this work, we propose DARE, a generative diffusion-based approach for autonomous exploration.
Given a partial map of the environment, DARE is trained to plan an exploration path that can extend into unexplored areas. From ground truth path demonstrations, DARE learns to reason about unexplored areas and plan efficient exploration paths. In doing so, we believe that planned explicit long-term paths provide a degree of explainability for learned models.
In our tests, DARE achieves on-par performance with the state-of-the-art planner. Through high-fidelity ROS simulations and hardware validation, we validate DARE's ability to operate in real time and bridge the sim-to-real gap.

Future works will focus on accelerating the inference speed of the diffusion policy to allow the planner to be more reactive to high-frequency map updates. We are also interested in further pushing the capacity of diffusion models for robot exploration, such as training diffusion models to explicitly predict the structure of unknown areas and leverage them for better exploration path planning.






\bibliographystyle{IEEEtran}
\bibliography{ref.bib}

\end{document}